\documentclass[11pt,a4paper]{article}
\usepackage[hyperref]{naaclhlt2019}
\usepackage{times}
\usepackage{latexsym}
\usepackage{graphicx}
\usepackage{url}
\usepackage{booktabs}

\aclfinalcopy

\title{SemEval-2019 Task 6: Identifying and Categorizing Offensive Language in Social Media (OffensEval)}

\author{Marcos Zampieri,\textsuperscript{1} Shervin Malmasi,\textsuperscript{2} Preslav Nakov,\textsuperscript{3}\\ \bf{Sara Rosenthal,\textsuperscript{4} Noura Farra,\textsuperscript{5} Ritesh Kumar\textsuperscript{6}} \\
  \textsuperscript{1}University of Wolverhampton, UK, \textsuperscript{2}Amazon Research, USA\\ 
  \textsuperscript{3}Qatar Computing Research Institute, HBKU, Qatar \\ 
  \textsuperscript{4}IBM Research, USA,  \textsuperscript{5}Columbia University, USA, \textsuperscript{6}Bhim Rao Ambedkar University, India \\
  {\tt m.zampieri@wlv.ac.uk} \\}

\date{}

\begin{document}
\maketitle
\begin{abstract}
 We present the results and the main findings of SemEval-2019 Task 6 on Identifying and Categorizing Offensive Language in Social Media (OffensEval). The task was based on a new dataset, the Offensive Language Identification Dataset (OLID), which contains over 14,000 English tweets. It featured three sub-tasks. In sub-task A, the goal was to discriminate between offensive and non-offensive posts. In sub-task B, the focus was on the type of offensive content in the post. Finally, in sub-task C, systems had to detect the target of the offensive posts. OffensEval attracted a large number of participants and it was one of the most popular tasks in SemEval-2019. In total, about 800 teams signed up to participate in the task, and 115 of them submitted results, which we present and analyze in this report.
\end{abstract}

\section{Introduction}

Recent years have seen the proliferation of offensive language in social media platforms such as Facebook and Twitter. As manual filtering is very time consuming, and as it can cause post-traumatic stress disorder-like symptoms to human annotators, there have been many research efforts aiming at automating the process. The task is usually modeled as a supervised classification problem, where systems are trained on posts annotated with respect to the presence of some form of abusive or offensive content. Examples of offensive content studied in previous work include hate speech \cite{davidson2017automated,malmasi2017,malmasi2018}, cyberbulling \cite{dinakar2011modeling}, and aggression \cite{kumar2018benchmarking}.
Moreover, given the multitude of terms and definitions used in the literature, some recent studies have investigated the common aspects of different abusive language detection sub-tasks \cite{waseem2017understanding,wiegand2018overview}. 

\noindent Interestingly, none of this previous work has studied both the type \textit{and} the target of the offensive language, which is our approach here.
Our task, OffensEval\footnote{\url{http://competitions.codalab.org/competitions/20011}}, uses the Offensive Language Identification Dataset (OLID)\footnote{\url{http://scholar.harvard.edu/malmasi/olid}} \cite{OLID}, which we created specifically for this task. OLID is annotated following a hierarchical three-level annotation schema that takes both the target and the type of offensive content into account.
Thus, it can relate to phenomena captured by previous datasets such as the one by \newcite{davidson2017automated}. Hate speech, for example, is commonly understood as an insult targeted at a {\em group}, whereas cyberbulling is typically targeted at an {\em individual}. 

We defined three sub-tasks, corresponding to the three levels in our annotation schema:\footnote{A total of 800 teams signed up to participate in the task, but only 115 teams ended up submitting results eventually.}

\begin{description}
\item[Sub-task A:] {Offensive language identification (104 participating teams)}
\item[Sub-task B:] {Automatic categorization of offense types (71 participating teams)}
\item[Sub-task C:] {Offense target identification (66 participating teams)}
\end{description}

The remainder of this paper is organized as follows: Section~\ref{sec:related} discusses prior work, including shared tasks related to OffensEval. Section \ref{sec:description} presents the shared task description and the sub-tasks included in OffensEval. Section \ref{sec:data} includes a brief description of OLID based on \cite{OLID}. Section \ref{sec:results} discusses the participating systems and their results in the shared task. Finally, Section \ref{sec:conclusion} concludes and suggests directions for future work.

\section{Related Work}
\label{sec:related}

Different abusive and offense language identification problems have been explored in the literature ranging from  aggression to cyber bullying, hate speech, toxic comments, and offensive language. Below we discuss each of them briefly.

\begin{table*}[t]
\centering
\begin{tabular}{p{12cm}ccc}
\toprule
\bf Tweet & \bf A & \bf B & \bf C \\
\midrule
 @USER He is so generous with his offers. & NOT & --- &  --- \\
 IM FREEEEE!!!! WORST EXPERIENCE OF MY FUCKING LIFE & OFF & UNT & --- \\
 @USER Fuk this fat cock sucker & OFF & TIN & IND \\
 @USER Figures! What is wrong with these idiots? Thank God for @USER & OFF & TIN & GRP \\
\bottomrule
\end{tabular}
\caption{Four tweets from the OLID dataset, with their labels for each level of the annotation model.}
\label{tab:examples}
\end{table*}

\vspace{3mm}

\noindent {\bf Aggression identification:} The TRAC shared task on Aggression Identification \cite{kumar2018benchmarking} provided participants with a dataset containing 15,000 annotated Facebook posts and comments in English and Hindi for training and validation. For testing, two different sets, one from Facebook and one from Twitter, were used. The goal was to discriminate between three classes: non-aggressive, covertly aggressive, and overtly aggressive. 
The best-performing systems in this competition used deep learning approaches based on convolutional neural networks (CNN), recurrent neural networks, and LSTM  \cite{aroyehun2018aggression,majumder2018filtering}. 

\vspace{3mm}

\noindent {\bf Bullying detection:} There have been several studies on cyber bullying detection. For example, \newcite{xu2012learning} used sentiment analysis and topic models to identify relevant topics, and \newcite{dadvar2013improving} used user-related features such as the frequency of profanity in previous messages.

\vspace{3mm}

\noindent {\bf Hate speech identification:} This is the most studied abusive language detection task \cite{kwok2013locate,burnap2015cyber,djuric2015hate}.
More recently, \newcite{davidson2017automated} presented the hate speech detection dataset with over 24,000 English tweets labeled as non offensive, hate speech, and profanity.

\vspace{3mm}

\noindent {\bf Offensive language:} The GermEval\footnote{\url{http://projects.fzai.h-da.de/iggsa/}} \cite{wiegand2018overview} shared task focused on offensive language identification in German tweets. A dataset of over 8,500 annotated tweets was provided for a course-grained binary classification task in which systems were trained to discriminate between offensive and non-offensive tweets. There was also a second task where 
the offensive class was subdivided into profanity, insult, and abuse. 
This is similar to our work, but there are three key differences: (\emph{i})~we have a third level in our hierarchy, (\emph{ii})~we use different labels in the second level, and (\emph{iii})~we focus on English.

\vspace{3mm}

\noindent {\bf Toxic comments:} The Toxic Comment Classification Challenge\footnote{\tiny\url{http://kaggle.com/c/jigsaw-toxic-comment-classification-challenge}} 
was an open competition at Kaggle, which provided participants with comments from Wikipedia organized in six classes: toxic, severe toxic, obscene, threat, insult, identity hate. The dataset was also used outside of the competition \cite{georgakopoulos2018convolutional}, including as additional training material for the aforementioned TRAC shared \cite{fortuna2018merging}.

\vspace{3mm}

While each of the above tasks tackles a particular type of abuse or offense, there are many commonalities. 
For example, an insult targeted at an individual is commonly known as cyberbulling and insults targeted at a group are known as hate speech.
The hierarchical annotation model proposed in OLID \cite{OLID} and used in OffensEval aims to capture this.  
We hope that the OLID's dataset would become a useful resource for various offensive language identification tasks.

\section{Task Description and Evaluation}
\label{sec:description}

The training and testing material for OffensEval is the aforementioned Offensive Language Identification Dataset (OLID) dataset, which was built specifically for this task. OLID was annotated using a hierarchical three-level annotation model introduced in \newcite{OLID}. Four examples of annotated instances from the dataset are presented in Table \ref{tab:examples}. We use the annotation of each of the three layers in OLID for a sub-task in OffensEval as described below. 

\subsection{Sub-task A: Offensive language identification}

In this sub-task, the goal is to discriminate between offensive and non-offensive posts. Offensive posts include insults, threats, and posts containing any form of untargeted profanity. Each instance is assigned one of the following two labels.

\begin{itemize}
\item Not Offensive (NOT): Posts that do not contain offense or profanity; 
\item Offensive (OFF): We label a post as offensive if it contains any form of non-acceptable language (profanity) or a targeted offense, which can be veiled or direct. This category includes insults, threats, and posts containing profane language or swear words.
\end{itemize}

\subsection{Sub-task B: Automatic categorization of offense types}

In sub-task B, the goal is to predict the type of offense. Only posts labeled as Offensive (OFF) in sub-task A are included in sub-task B. The two categories in sub-task B are the following:

\begin{itemize}

\item Targeted Insult (TIN): Posts containing an insult/threat to an individual, group, or others (see sub-task C below);

\item Untargeted (UNT): Posts containing non-targeted profanity and swearing. Posts with general profanity are not targeted, but they contain non-acceptable language.

\end{itemize}

\subsection{Sub-task C: Offense target identification}

Sub-task C focuses on the target of offenses. Only posts that are either insults or threats (TIN) arwe considered in this third layer of annotation. The three labels in sub-task C are the following:

\begin{itemize}

\item Individual (IND): Posts targeting an individual. It can be a a famous person, a named individual or an unnamed participant in the conversation. Insults/threats targeted at individuals are often defined as cyberbullying. 

\item Group (GRP): The target of these offensive posts is a group of people considered as a unity due to the same ethnicity, gender or sexual orientation, political affiliation, religious belief, or other common characteristic. Many of the insults and threats targeted at a group correspond to what is commonly understood as hate speech.

\item Other (OTH): The target of these offensive posts does not belong to any of the previous two categories, e.g.,~an organization, a situation, an event, or an issue. 

\end{itemize}

\subsection{Task Evaluation}

Given the strong imbalance between the number of instances in the different classes across the three tasks, we used the macro-averaged F1-score as the official evaluation measure for all three sub-tasks.

\begin{figure}
    \centering
    \includegraphics[width=\columnwidth]{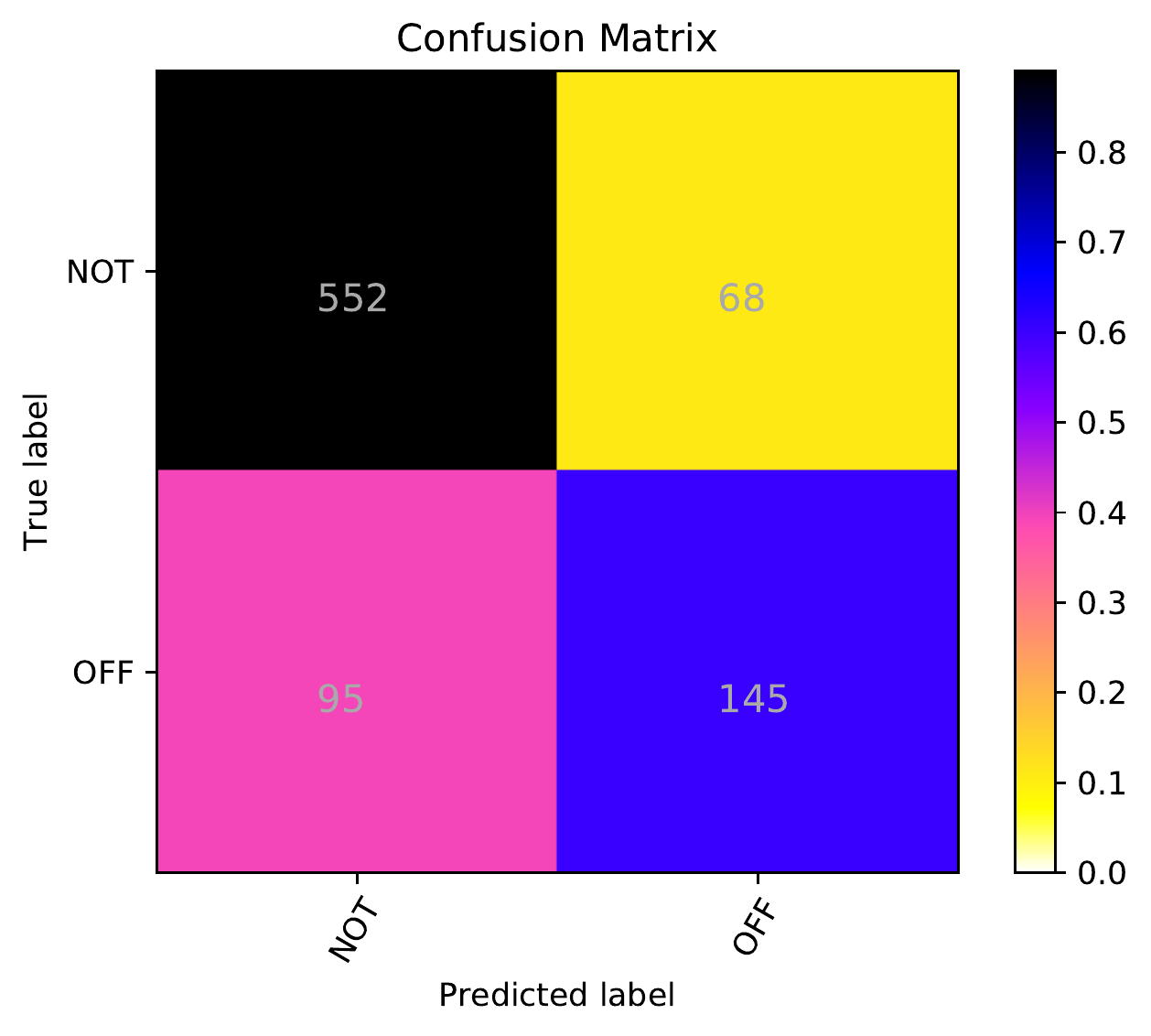}
\caption{Example of a confusion matrix provided in the results package for team NULI, which is the best-performing team for sub-task A.}
    \label{F:confmatrix}
\end{figure}

At the end of the competition, we provided the participants with packages containing the results for each of their submissions, including tables and confusion matrices, and tables with the ranks listing all teams who competed in each sub-task. For example, the confusion matrix for the best team in sub-task A is shown in Figure~\ref{F:confmatrix}.

\subsection{Participation}

\begin{table}[ht!]
\setlength{\tabcolsep}{1.1pt}
\scalebox{0.92}{
\begin{tabular}{ll}
   {\bf Team}  & {\bf System Description Paper} \\ 
\midrule
Amobee & \cite{alonrozental} \\
ASE-CSE & \cite{amrita} \\
bhanodaig & \cite{Riteshkumar} \\
\footnotesize{BNU-HKBU ...} & \cite{Wu2019UIC} \\
CAMsterdam & \cite{caines} \\
CN-HIT-MI.T & \cite{Zhang} \\
ConvAI & \cite{Pavlopoulos}\\
DA-LD-Hildesheim & \cite{Modha} \\
DeepAnalyzer & \cite{deepanalyser} \\
Duluth & \cite{Pedersen} \\
Emad & \cite{emad} \\
Embeddia & \cite{pelicon} \\
Fermi & \cite{fermi} \\
Ghmerti & \footnotesize{\cite{doostmohammadi}} \\ 
HAD-T{\"u}bingen & \cite{Bansal} \\
HHU & \cite{romberg} \\
Hope & \cite{hope} \\
INGEOTEC & \cite{Graff} \\
JCTICOL & \footnotesize{\cite{HaCohen-Kerner}} \\
jhan014 & \cite{han} \\
JTML & \cite{jtml} \\
JU\_ETCE\_17\_21 & \cite{mukherjeeetal} \\
KMI\_Coling & \cite{kmi} \\
LaSTUS/TALN & \cite{Mut} \\
LTL-UDE & \cite{aggarwal} \\
MIDAS & \cite{midas} \\
Nikolov-Radivchev & \cite{Nikolov} \\
\footnotesize{NIT\_Agartala\_NLP\_Team} & \cite{swamy} \\
NLP & \cite{asif} \\
NLP@UIOWA & \cite{rusert} \\
NLPR@SRPOL & \cite{Seganti} \\
nlpUP & \cite{mitrovic} \\
NULI & \cite{nuli} \\
SINAI & \footnotesize{\cite{Plaza-del-Arco}} \\
SSN\_NLP & \cite{thenmozhi} \\
Stop PropagHate & \cite{fortunaSemEval} \\
Pardeep & \cite{Singh} \\
techssn & \cite{techssn} \\
The Titans & \cite{garain} \\
TUVD & \cite{shushkevich} \\
T{\"u}KaSt & \cite{Kannan} \\
UBC-NLP & \cite{Rajendran} \\
UTFPR  & \cite{paetzold}   \\
UHH-LT & \cite{Wiedemann} \\
UM-IU@LING & \cite{zhu} \\
USF & \cite{Goel} \\
UVA Wahoos & \cite{ramakrishnan} \\
YNU-HPCC & \cite{zhou} \\
YNUWB & \cite{Wang} \\
Zeyad & \cite{elzanaty} \\
\bottomrule
    \end{tabular}
}
\caption{The teams that participated in OffensEval and submitted system description papers.}
\label{tab:teams}
\end{table}

The task attracted nearly 800 teams and 115 of them submitted their results. The teams that submitted papers for the SemEval-2019 proceedings are listed in Table \ref{tab:teams}.\footnote{\emph{ASE-CSE} is for \emph{Amrita School of Engineering - CSE}.}

\section{Data}
\label{sec:data}

Below, we briefly describe OLID, the dataset used for our SemEval-2019 task 6. A detailed description of the data collection process and annotation is presented in \newcite{OLID}.

OLID is a large collection of English tweets annotated using a hierarchical three-layer annotation model. It contains 14,100 annotated tweets divided into a training partition of 13,240 tweets and a testing partition of 860 tweets. Additionally, a small trial dataset of 320 tweets was made available before the start of the competition. 

\begin{table}[!ht]
\centering
\begin{tabular}{cccrrr}
\toprule
\bf A & \bf B &  \bf C & \bf Train & \bf Test & \bf Total \\ 
\midrule
OFF  &  TIN  &  IND &  $2{,}407$ & $100$ & $2{,}507$\\
OFF  &  TIN  &  OTH &  $395$ & $35$ & $430$ \\
OFF  &  TIN  &  GRP &  $1{,}074$ & $78$ & $1{,}152$\\ 
OFF  &  UNT  &  --- &  $524$ & $27$ & $551$ \\
NOT  &  ---  &  --- &  $8{,}840$ & $620$ & $9{,}460$ \\
\midrule
\bf All &    &     &   $13{,}240$ & $860$ & $14{,}100$ \\
\bottomrule 
\end{tabular}
\caption{Distribution of label combinations in OLID.}
\label{tab:labels}
\end{table}

The distribution of the labels in OLID is shown in Table~\ref{tab:labels}.
We annotated the dataset using the crowdsourcing platform Figure Eight.\footnote{\url{https://www.figure-eight.com/}} We ensured the quality of the annotation by only hiring experienced annotators on the platform and by using test questions to discard annotators who did not achieve a certain threshold. All the tweets were annotated by two people. In case of disagreement, a third annotation was requested, and ultimately we used a majority vote. Examples of tweets from the dataset with their annotation labels are shown in Table \ref{tab:examples}.

\begin{figure}
    \centering
    \includegraphics[width=\columnwidth]{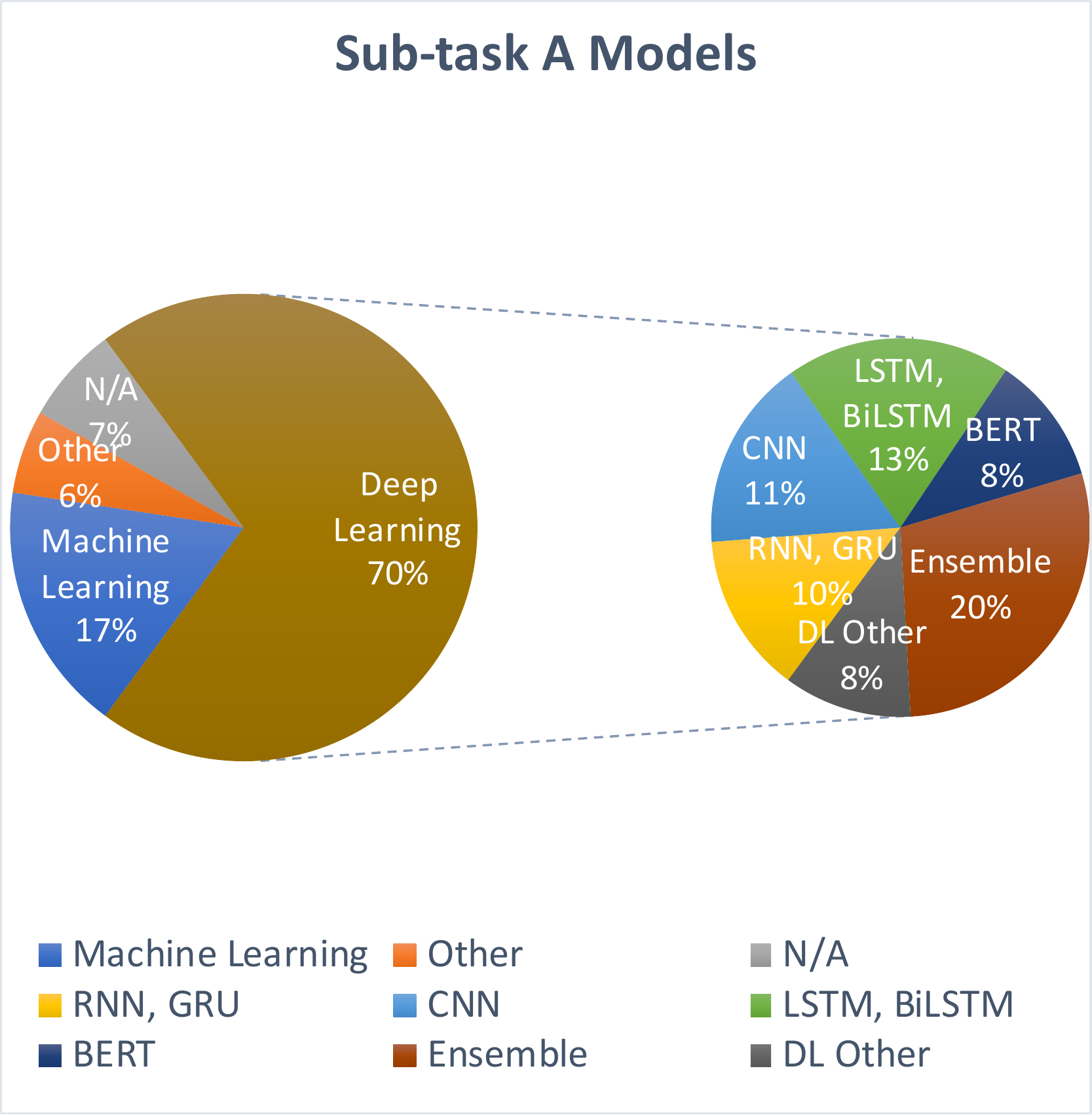}
\caption{Pie chart showing the models used in sub-task A. `N/A' indicates that the system did not have a description.}
    \label{F:A-Models}
\end{figure}

\section{Results}
\label{sec:results}

The models used in the task submissions ranged from traditional machine learning, e.g., SVM and logistic regression, to deep learning, e.g., CNN, RNN, BiLSTM, including attention mechanism, to state-of-the-art deep learning models such as ELMo~\cite{Peters:2018} and BERT~\cite{DBLP:journals/corr/abs-1810-04805}. Figure~\ref{F:A-Models} shows a pie chart indicating the breakdown by model type for all participating systems in sub-task A. Deep learning was clearly the most popular approach, as were also ensemble models. Similar trends were observed for sub-tasks B and C.

\noindent Some teams used additional training data, exploring external datasets such as Hate Speech Tweets \cite{davidson2017automated}, toxicity labels \cite{Thain2017}, and TRAC \cite{kumar2018benchmarking}. Moreover, seven teams indicated that they used sentiment lexicons or a sentiment analysis model for prediction, and two teams reported the use of offensive word lists. Furthermore, several teams used pre-trained word embeddings from FastText \cite{DBLP:journals/corr/BojanowskiGJM16}, from GloVe, including Twitter embeddings from GloVe \cite{pennington2014glove} and from word2vec~\cite{Mikolov:2013:DRW:2999792.2999959,W15-4322}.

In addition, several teams used techniques for pre-processing the tweets such as normalizing the tokens, hashtags, URLs, retweets (RT), dates, elongated words (e.g., ``Hiiiii'' to ``Hi'', partially hidden words (``c00l'' to ``cool''). Other techniques include converting emojis to text, removing uncommon words, and using Twitter-specific tokenizers, such as the Ark Tokenizer\footnote{\url{http://www.cs.cmu.edu/~ark/TweetNLP}}~\cite{Gimpel:2011:PTT:2002736.2002747} and the NLTK TweetTokenizer,\footnote{\url{http://www.nltk.org/api/nltk.tokenize.html}} as well as standard tokenizers (Stanford Core NLP~\cite{manning-EtAl:2014:P14-5}, and the one from Keras.\footnote{\url{http://keras.io/preprocessing/text/}} Approximately a third of the teams indicated that they used one or more of these techniques.

\noindent The results for each of the sub-tasks are shown in Table~\ref{T:results}. Due to the large number of submissions, we only show the F1-score for the top-10 teams, followed by result ranges for the rest of the teams. We further include the models and the baselines from \cite{OLID}: CNN, BiLSTM, and SVM. The baselines are choosing all predictions to be of the same class, e.g., all offensive, and all not offensive for sub-task A. Table~\ref{T:teams} shows all the teams that participated in the tasks along with their ranks in each task. These two tables can be used together to find the score/range for a particular team. \\

Below, we describe the overall results for each sub-task, and we describe the top-3 systems.

\subsection{Sub-task A}

Sub-task A was the most popular sub-task with 104 participating teams. Among the top-10 teams, seven used BERT~\cite{DBLP:journals/corr/abs-1810-04805} with variations in the parameters and in the pre-processing steps. The top-performing team, \textit{NULI}, used BERT-base-uncased with default-parameters, but with a max sentence length of 64 and trained for 2 epochs. The 82.9\% F1 score of NULI is 1.4 points better than the next system, but the difference between the next 5 systems, ranked 2-6, is less than one point: 81.5\%-80.6\%. The top non-BERT model, \textit{MIDAS}, is ranked sixth. They used an ensemble of CNN and BLSTM+BGRU, together with Twitter word2vec embeddings \cite{W15-4322} and token/hashtag normalization.

\begin{table*}[!ht]
\centering
\begin{tabular}{cc|cc|cc}
\toprule
\multicolumn{2}{c}{\bf Sub-task A} & \multicolumn{2}{c}{\bf Sub-task B} & \multicolumn{2}{c}{\bf Sub-task C} \\
\bf Team Ranks	&	\bf F1 Range	&	\bf Team Ranks	&	\bf F1 Range	&	\bf Team Ranks	&	\bf F1 Range	\\
\midrule
1	&	0.829	&	1	&	0.755	&	1	&	0.660	\\
2	&	0.815	&	2	&	0.739	&	2	&	0.628	\\
3	&	0.814	&	3	&	0.719	&	3	&	0.626	\\
4	&	0.808	&	4	&	0.716	&	4	&	0.621	\\
5	&	0.807	&	5	&	0.708	&	5	&	0.613	\\
6	&	0.806	&	6	&	0.706	&	6	&	0.613	\\
7	&	0.804	&	7	&	0.700	&	7	&	0.591	\\
8	&	0.803	&	8	&	0.695	&	8	&	0.588	\\
9	&	0.802	&	9	&	0.692	&	9	&	0.587	\\
\textbf{CNN}	&	\textbf{0.800}	&	\textbf{CNN}	&	\textbf{0.690}	&	10	&	0.586	\\
10	&	0.798	&	10	&	0.687	&	11-14	&	.571-.580	\\
11-12	&	.793-.794	&	11-14	&	.680-.682	&	15-18	&	.560-.569	\\
13-23	&	.782-.789	&	15-24	&	.660-.671	&	19-23	&	.547-.557	\\
24-27	&	.772-.779	&	\textbf{BiLSTM}	&	\textbf{0.660}	&	24-29	&	.523-.535	\\
28-31	&	.765-.768	&	25-29	&	.640-.655	&	30-33	&	.511-.515	\\
32-40	&	.750-.759	&	\textbf{SVM}	&	\textbf{0.640}	&	34-40	&	.500-.509	\\
\textbf{BiLSTM}	&	\textbf{0.750}	&	30-38	&	.600-.638	&	41-47	&	.480-.490	\\
41-45	&	.740-.749	&	39-49	&	.553-.595	&	\textbf{CNN}	&\textbf{0.470}	\\
46-57	&	.730-.739	&	50-62	&	.500-.546	&	\textbf{BiLSTM}	&	\textbf{0.470}	\\
58-63	&	.721-.729	&	\textbf{ALL TIN}	&	\textbf{0.470}	&	\textbf{SVM}	&	\textbf{0.450}	\\
64-71	&	.713-.719	&	63-74	&	.418-.486	&	46-60	&	.401-.476	\\
72-74	&	.704-.709	&	75	&	0.270	&	61-65	&	.249-.340	\\
\textbf{SVM}	&	\textbf{0.690}	&	76	&	0.121	&	\textbf{All IND}	&	\textbf{0.210}	\\
75-89	&	.619-.699	&	\textbf{All UNT}	&	\textbf{0.100}	&	\textbf{All GRP	}&	\textbf{0.180}	\\
90-96	&	.500-.590	&		&		&	\textbf{ALL OTH}	&	\textbf{0.090}	\\
97-103	&	.422-.492	&		&		&		&		\\
\textbf{All NOT}	&	\textbf{0.420}	&		&		&		&		\\
\textbf{All OFF}	&	\textbf{0.220}	&		&		&		&		\\
104	&	0.171	&		&		&		&		\\
\bottomrule
\end{tabular}
\caption{F1-Macro for the top-10 teams followed by the rest of the teams grouped in ranges for all three sub-tasks. Refer to Table~\ref{T:teams} to see the team names associated with each rank. We also include the models (\textbf{CNN, BiLSTM, and SVM}) and the baselines (\textbf{All NOT and All OFF}) from \cite{OLID}, shown in bold.}
\label{T:results}
\end{table*}

\subsection{Sub-task B}

A total of 76 teams participated in sub-task B, and 71 of them had also participated in sub-task A. In contrast to sub-task A, where BERT clearly dominated, here five of the top-10 teams used an ensemble model. Interestingly, the best team, \textit{jhan014}, which was ranked 76th in sub-task A, used a rule-based approach with a keyword filter based on a Twitter language behavior list, which included strings such as hashtags, \@ signs, etc., achieving an F1-score of 75.5\%. The second and the third teams, 
\textit{Amobee} and \textit{HHU}, used ensembles of deep learning (including BERT) and non-neural machine learning models. The best team from sub-task A also performed well here, ranked 4th (71.6\%), thus indicating that overall BERT works well for sub-task B as well.

\subsection{Sub-task C}

A total of 66 teams participated in sub-task C, and most of them also participated in sub-tasks A and B. As in sub-task B, ensembles were quite successful and were used by five of the top-10 teams. However, as in sub-task A, the best team, \textit{vradivchev\_anikolov}, used BERT after trying many other deep learning methods. They also used pre-processing and pre-trained word embeddings based on GloVe. The second best team, \textit{NLPR$@$SRPOL}, used an ensemble of deep learning models such as OpenAI Finetune, LSTM, Transformer, and non-neural machine learning models such as SVM and Random Forest.

\subsection{Description of the Top Teams} 

The top-3 teams by average rank for all three sub-tasks were \textit{NLPR$@$SRPOL}, \textit{NULI}, and \textit{vradivchev\_anikolov}. Below, we provide a brief description of their approaches:

\begin{table*}[h!]
\setlength{\tabcolsep}{1pt}
\centering
\begin{tabular}{lccc@{ }@{ }@{ }@{ }@{ }@{ }lccc@{ }@{ }@{ }@{ }@{ }@{ }lccc}
\\
\\
\\
\\
\toprule
& \multicolumn{3}{c}{\bf Sub-task} & & \multicolumn{3}{c}{\bf Sub-task} & & \multicolumn{3}{c}{\bf Sub-task} \\
\bf Team	&	\bf A	&	\bf B	&	\bf C	&	\bf Team	&	\bf A	&	\bf B	&	\bf C	&	\bf Team	&	\bf A	&	\bf B	&	\bf C	\\
\midrule
\textbf{NULI}	&	\textbf{1}	&	4	&	18	&	resham	&	40	&	43	&	-	&	kroniker	&	79	&	71	&	-	\\
\underline{\textbf{vradivchev\_anikolov}}	&	\underline{2}	&	16	&	\textbf{1}	&	Xcosmos	&	41	&	47	&	29	&	aswathyprem	&	80	&	-	&	-	\\
UM-IU$@$LING	&	3	&	76	&	27	&	jkolis	&	42	&	-	&	-	&	DeepAnalyzer	&	81	&	38	&	45	\\
Embeddia	&	4	&	18	&	5	&	\footnotesize{NIT\_Agartala\_NLP\_Team}	&	43	&	5	&	38	&	Code Lyoko	&	82	&	-	&	-	\\
MIDAS	&	5	&	8	&	-	&	Stop PropagHate	&	44	&	-	&	-	&	rowantahseen	&	83	&	-	&	-	\\
BNU-HKBU &	6	&	62	&	39	&	KVETHZ	&	45	&	52	&	26	&	ramjib	&	84	&	-	&	-	\\
SentiBERT	&	7	&	-	&	-	&	christoph.alt	&	46	&	14	&	36	&	OmerElshrief	&	85	&	-	&	-	\\
\underline{NLPR$@$SRPOL}	&	8	&	9	&	\underline{2}	&	TECHSSN	&	47	&	22	&	16	&	desi	&	86	&	56	&	-	\\
YNUWB	&	9	&	-	&	-	&	USF	&	48	&	32	&	62	&	Fermi	&	87	&	31	&	3	\\
LTL-UDE	&	10	&	-	&	19	&	Ziv\_Ben\_David	&	49	&	64	&	33	&	mkannan	&	88	&	-	&	-	\\
nlpUP	&	11	&	-	&	-	&	JCTICOL	&	50	&	63	&	-	&	mking	&	89	&	35	&	54	\\
ConvAI	&	12	&	11	&	35	&	T{\"u}KaSt	&	51	&	23	&	50	&	ninab	&	90	&	69	&	-	\\
Vadym	&	13	&	10	&	-	&	Gal\_DD	&	52	&	66	&	25	&	dianalungu725	&	91	&	74	&	65	\\
UHH-LT	&	14	&	21	&	13	&	HAD-T{\"u}bingen	&	53	&	59	&	61	&	Halamulki	&	92	&	-	&	-	\\
CAMsterdam	&	15	&	19	&	20	&	Emad	&	54	&	-	&	-	&	SSN\_NLP	&	93	&	65	&	64	\\
YNU-HPCC	&	16	&	-	&	-	&	NLP$@$UIOWA	&	55	&	27	&	37	&	UTFPR	&	94	&	-	&	-	\\
nishnik	&	17	&	-	&	-	&	INGEOTEC	&	56	&	15	&	12	&	rogersdepelle	&	95	&	-	&	-	\\
\underline{Amobee}	&	18	&	\underline{2}	&	7	&	Duluth	&	57	&	39	&	44	&	Amimul Ihsan	&	96	&	-	&	-	\\
himanisoni	&	19	&	46	&	11	&	Zeyad	&	58	&	34	&	34	&	supriyamandal	&	97	&	75	&	-	\\
samsam	&	20	&	-	&	-	&	ShalomRochman	&	59	&	70	&	58	&	ramitpahwa	&	98	&	-	&	-	\\
JU\_ETCE\_17\_21	&	21	&	50	&	47	&	stefaniehegele	&	60	&	-	&	-	&	ASE - CSE	&	99	&	33	&	32	\\
DA-LD-Hildesheim	&	22	&	28	&	21	&	NLP-CIC	&	61	&	48	&	46	&	kripo	&	100	&	-	&	-	\\
YNU-HPCC	&	23	&	12	&	4	&	Elyash	&	62	&	67	&	40	&	garain	&	101	&	44	&	63	\\
ChenXiuling	&	24	&	-	&	28	&	KMI\_Coling	&	63	&	45	&	53	&	NAYEL	&	102	&	-	&	-	\\
Ghmerti	&	25	&	29	&	-	&	RUG\_OffenseEval	&	64	&	-	&	-	&	magnito60	&	103	&	-	&	-	\\
safina	&	26	&	-	&	-	&	jaypee1996	&	65	&	41	&	-	&	AyushS	&	104	&	36	&	48	\\
Arjun Roy	&	27	&	17	&	-	&	orabia	&	66	&	55	&	8	&	UBC\_NLP	&	-	&	6	&	9	\\
CN-HIT-MI.T	&	28	&	30	&	22	&	v.gambhir15	&	67	&	58	&	60	&	bhanodaig	&	-	&	57	&	-	\\
LaSTUS/TALN	&	29	&	20	&	15	&	kerner-jct.ac.il	&	68	&	68	&	42	&	Panaetius	&	-	&	60	&	-	\\
HHU	&	30	&	3	&	-	&	SINAI	&	69	&	-	&	-	&	eruppert	&	-	&	61	&	-	\\
na14	&	31	&	26	&	10	&	apalmer	&	70	&	13	&	55	&	Macporal	&	-	&	72	&	-	\\
NRC	&	32	&	37	&	24	&	ayman	&	71	&	53	&	57	&	NoOffense	&	-	&	-	&	6	\\
NLP	&	33	&	54	&	52	&	Geetika	&	72	&	24	&	-	&	HHU	&	-	&	-	&	14	\\
JTML	&	34	&	-	&	-	&	Taha	&	73	&	51	&	59	&	quanzhi	&	-	&	-	&	17	\\
Arup-Baruah	&	35	&	25	&	31	&	justhalf	&	74	&	-	&	-	&	TUVD	&	-	&	-	&	23	\\
UVA\_Wahoos	&	36	&	42	&	-	&	Pardeep	&	75	&	7	&	41	&	mmfouad	&	-	&	-	&	51	\\
NLP$@$UniBuc	&	37	&	73	&	49	&	\textbf{jhan014}	&	76	&	\textbf{1}	&	30	&	balangheorghe	&	-	&	-	&	56	\\
NTUA-ISLab	&	38	&	40	&	43	&	liuxy94	&	77	&	-	&	-	&		&		&		&		\\
Rohit	&	39	&	49	&	-	&	ngre1989	&	78	&	-	&	-	&		&		&		&		\\
\bottomrule
\end{tabular}
\caption{All the teams that participated in SemEval-2019 Task 6 with their ranks for each sub-task. The symbol `-' indicates that the team did not participate in some of the subtasks. Please, refer to Table~\ref{T:results} to see the scores based on a team's rank. The top team for each task is in \textbf{bold}, and the second-place team is \underline{underlined}. Note: \emph{ASE - CSE} stands for \emph{Amrita School of Engineering - CSE}, and \emph{BNU-HBKU} stands for \emph{BNU-HKBU UIC NLP Team 2}.}
\label{T:teams}
\end{table*}

\begin{description}

\item[NLPR$@$SRPOL] was ranked 8th, 9th, and 2nd on sub-tasks A, B, and C, respectively. They used ensembles of OpenAI GPT, Random Forest, the Transformer, Universal encoder, ELMo, and combined embeddings from fastText and custom ones. They trained their models on multiple publicly available offensive datasets, as well as on their own custom dataset annotated by linguists.

\item[NULI] was ranked 1st, 4th, and 18th on sub-tasks A, B, and C, respectively. They experimented with different models including linear models, LSTM, and pre-trained BERT with fine-tuning on the OLID dataset. Their final submissions for all three subtasks only used BERT, which performed best during development. They also used a number of pre-processing techniques such as hashtag segmentation and emoji substitution.

\item[vradivchev\_anikolov] was ranked 2nd, 16th, and 1st on sub-tasks A, B, and C, respectively. 
They trained a variety of models and combined them in ensembles, but their best submissions for sub-tasks A and C used BERT only, as the other models overfitted. For sub-task B, BERT did not perform as well, and they used soft voting classifiers. In all cases, they used pre-trained GloVe vectors and they also applied techniques to address the class imbalance in the training data.

\end{description}

\section{Conclusion}
\label{sec:conclusion}

We have described SemEval-2019 Task 6 on Identifying and Categorizing Offensive Language in Social Media (OffensEval). The task used OLID \cite{OLID}, a dataset of English tweets annotated for offensive language use, following a three-level hierarchical schema that considers (\emph{i})~whether a message is offensive or not (for sub-task A), (\emph{ii})~what is the type of the offensive message (for sub-task B), and (\emph{iii})~who is the target of the offensive message (for sub-task C). 

Overall, about 800 teams signed up for OffensEval, and 115 of them actually participated in at least one sub-task. The evaluation results have shown that the best systems used ensembles and state-of-the-art deep learning models such as BERT. Overall, both deep learning and traditional machine learning classifiers were widely used. More details about the indvididual systems can be found in their respective system description papers, which are published in the SemEval-2019 proceedings. A list with references to these publications can be found in Table \ref{tab:teams}; note, however, that only 50 of the 115 participating teams submitted a system description paper.

As is traditional for SemEval, we have made OLID publicly available to the research community beyond the SemEval competition, hoping to facilitate future research on this important topic. 

\noindent In fact, the OLID dataset and the SemEval-2019 Task 6 competition setup have already been used in teaching curricula in universities in UK and USA. For example, student competitions based on OffensEval using OLID have been organized as part of Natural Language Processing and Text Analytics courses in two universities in UK: Imperial College London and the University of Leeds. System papers describing some of the students' work are publicly accessible\footnote{\url{http://scholar.harvard.edu/malmasi/offenseval-student-systems}} and have also been made available on arXiv.org \cite{cambray2019bidirectional,frisiani2019combination,ong2019transforma,sapora2019,bogdan2019,uglow2019}. Similarly, a number of students in Linguistics and Computer Science at the University of Arizona in USA have been using OLID in their coursework.

In future work, we plan to increase the size of the OLID dataset, while addressing issues such as class imbalance and the small size for the test partition, particularly for sub-tasks B and C. We would also like to expand the dataset and the task to other languages.

\section*{Acknowledgments}

We would like to thank the SemEval-2019 organizers for hosting the OffensEval task and for replying promptly to all our inquires. We further thank the SemEval-2019 anonymous reviewers for the helpful suggestions and for the constructive feedback, which have helped us improve the text of this report.

We especially thank the SemEval-2019 Task 6 participants for their interest in the shared task, for their participation, and for their timely feedback, which have helped us make the shared task a success.

Finally, we would like to thank Lucia Specia from Imperial College London and Eric Atwell from the University of Leeds for hosting the OffensEval competition in their courses. We further thank the students who participated in these student competitions and especially those who wrote papers describing their systems. 

The research presented in this paper was partially supported by an ERAS fellowship, which was awarded to Marcos Zampieri by the University of Wolverhampton, UK.

\bibliography{naaclhlt2019,systempapers}
\bibliographystyle{acl_natbib}

\end{document}